\documentclass[10pt,twocolumn,letterpaper]{article}

\usepackage{wacv}
\usepackage{times}
\usepackage{epsfig}
\usepackage{graphicx}
\usepackage{amsmath}
\usepackage{amssymb}

\usepackage{subfigure}
\usepackage{stfloats}
\usepackage{caption}
\usepackage[accsupp]{axessibility}  

\def\wacvPaperID{312} 

\wacvapplicationstrack

\wacvfinalcopy 

\ifwacvfinal
\fi

\ifwacvfinal
\usepackage[breaklinks=true,bookmarks=false]{hyperref}
\else
\usepackage[pagebackref=true,breaklinks=true,colorlinks,bookmarks=false]{hyperref}
\fi

\begin{document}

\title{PointNeuron: 3D Neuron Reconstruction via Geometry and Topology Learning of Point Clouds}

\author{Runkai Zhao\\
University of Sydney\\
{\tt\small rzha9419@uni.sydney.edu.au}
\and
Heng Wang\\
University of Sydney\\
{\tt\small hwan9147@uni.sydney.edu.au}
\and
Chaoyi Zhang\\
University of Sydney\\
{\tt\small czha5168@uni.sydney.edu.au}
\and
Weidong Cai\\
University of Sydney\\
{\tt\small tom.cai@sydney.edu.au}
}

\maketitle
\thispagestyle{empty}

\begin{abstract}
Digital neuron reconstruction from 3D microscopy images is an essential technique for investigating brain connectomics and neuron morphology. Existing reconstruction frameworks use convolution-based segmentation networks to partition the neuron from noisy backgrounds before applying the tracing algorithm. The tracing results are sensitive to the raw image quality and segmentation accuracy. In this paper, we propose a novel framework for 3D neuron reconstruction. Our key idea is to use the geometric representation power of the point cloud to better explore the intrinsic structural information of neurons. Our proposed framework adopts one graph convolutional network to predict the neural skeleton points and another one to produce the connectivity of these points. We finally generate the target SWC file through interpretation of the predicted point coordinates, radius, and connections. Evaluated on the Janelia-Fly dataset from the BigNeuron project, we show that our framework achieves competitive neuron reconstruction performance. Our geometry and topology learning of point clouds could further benefit 3D medical image analysis, such as cardiac surface reconstruction. Our code is available at https://github.com/RunkaiZhao/PointNeuron.
\end{abstract}

\section{Introduction}

\begin{figure}[t]
\centering
\includegraphics[width=0.8\linewidth]{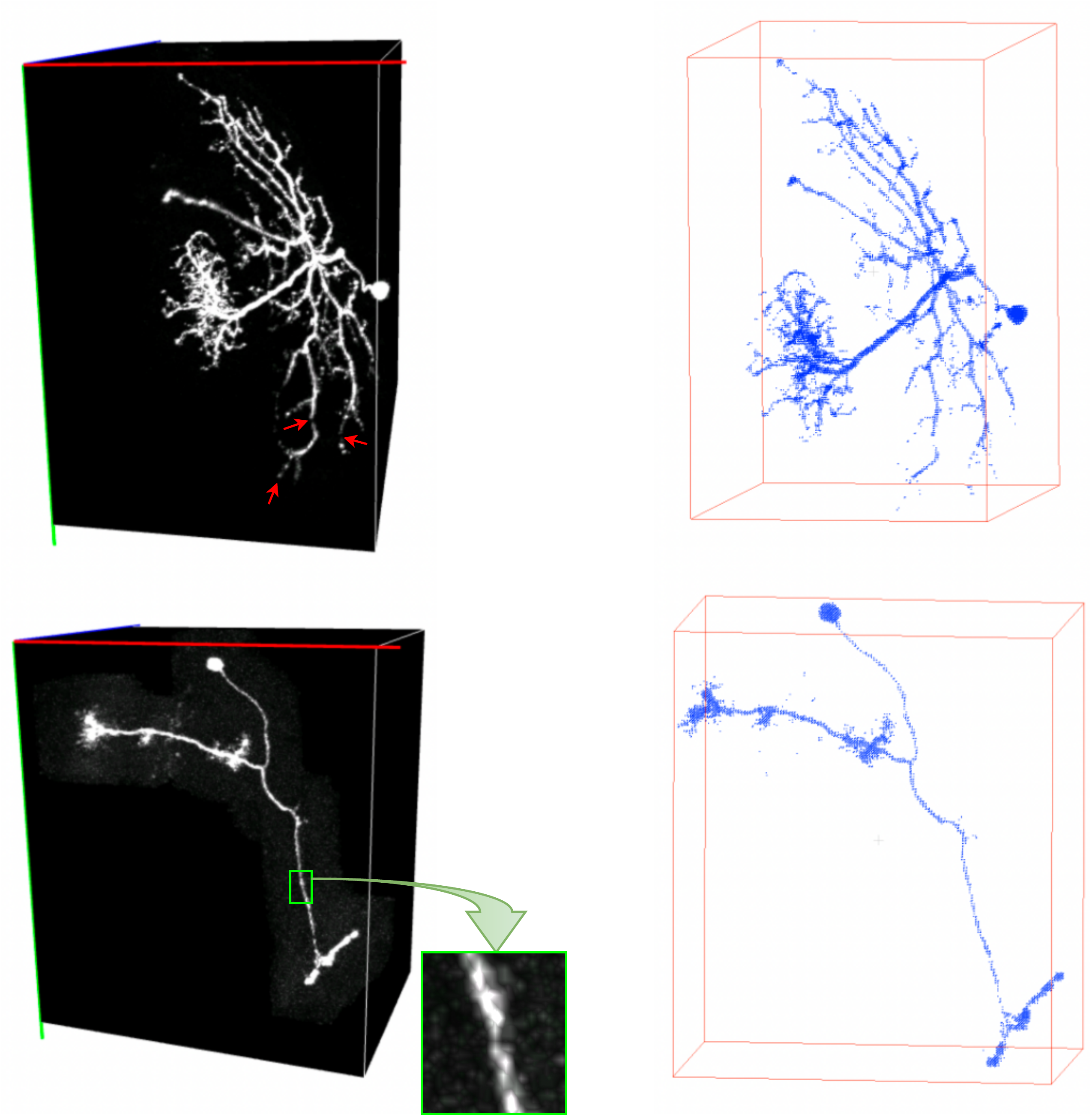}
\caption{We re-think the structural representation of neuron by using point cloud. Left: the voxel-wise neuron microscopy image; Right: the point-wise neuron after the format transformation. Due to the limits of optical microscope imaging, there are gaps along neuron tree-like arbors (red arrows), and the neuron structure is surrounded by the background noises (green box). Note voxel-based representation is inherently dense in three dimensions while our point-based one is more efficient in memory (e.g., $200\times100\times150$ voxels versus $4500$ points).}
\label{fig: img_to_point}
\end{figure}

Neuron morphology plays an essential role in the analysis of brain functionality. Digital 3D neuron reconstruction, also named 3D neuron tracing, is a computer-aided process to extract the anatomical structure and connectivity of neuron circuits from the volumetric microscopy image. 
Acquisitions of neuron morphology models in the past several decades have relied on manual annotation from neuroscientists. Due to the diversity and complexity of neuron morphology, the manual annotation work is extremely time-consuming and labor-intensive. The manual annotations are recorded as SWC files for digitally storing the neuron morphologies which use a set of connected points to constitute the hierarchical neuronal trees. It includes the identity of each neuron node, such as ID, type, position, radius, and parent ID. Recently, many researchers have devoted more attention to completing neuron reconstruction in an automatic or semi-automatic manner. The BigNeuron challenge \cite{peng2015bigneuron} and the DIADEM challenge \cite{brown2011diadem} have been hosted to develop automatic tracing algorithms by providing a sizeable single neuron morphology database and open-source software tools for neuroscience studies. 

Early digital neuron reconstruction algorithms relied on sophisticated mathematical models, which can be categorized into global and local algorithms. The global approaches, such as open-curve snake \cite{gloabl_wang2011broadly}, APP \cite{global-peng2011automatic}, APP2 \cite{global-xiao2013app2}, FMST \cite{global-yang2019fmst}, and others \cite{global_lee2012high, global_myatt2012neuromantic, global-turetken2013reconstructing, global-gala2014active, global-sui2014pipeline, wang2018memory}, consist of multiple stages which are pre-processing to denoise the raw image, tree-like structure initialization, and post-processing to refine the reconstructed traces. The local approaches \cite{local-zhao2011automated, local-bas2011principal, local-choromanska2012automatic} are to trace the neuronal tree from the seed point location with manual intervention or automatic detection.

\begin{figure*}[t]
    \centering
    \includegraphics[width=\linewidth]{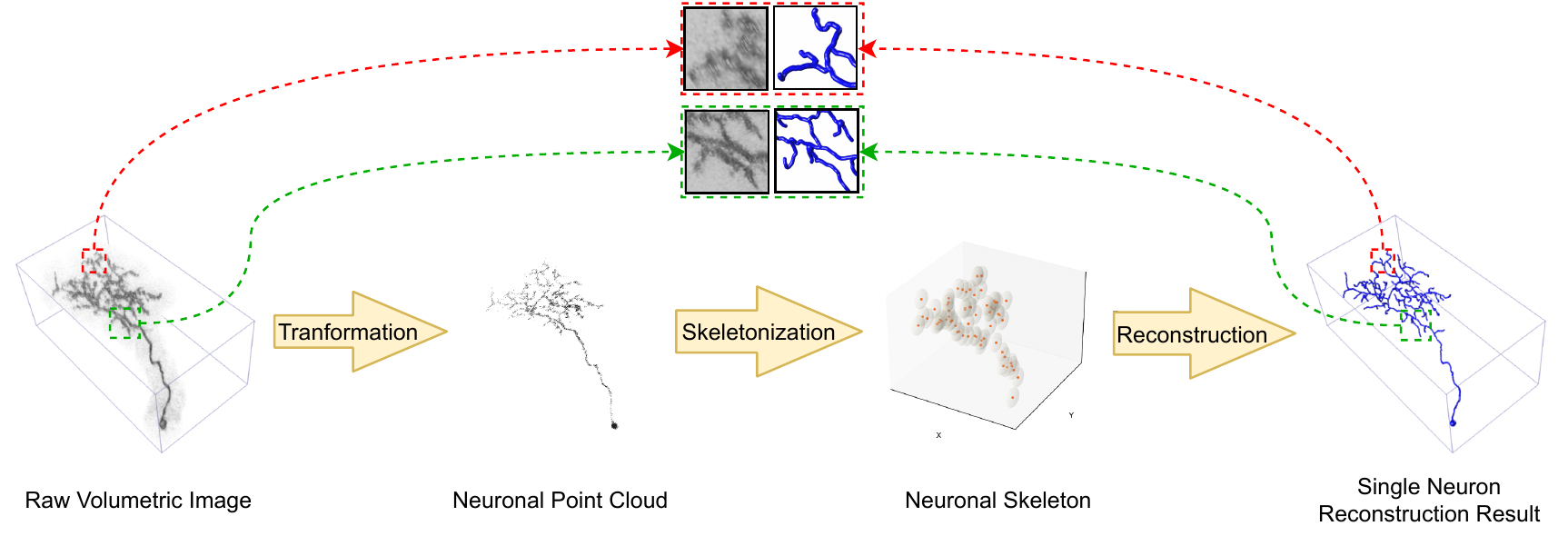}
    \caption{The main procedures of our proposed method, \emph{PointNeuron}, for neuron reconstuction from a volumetric microscopy image. The green and red boxes highlight parts of our reconstruction improvements.}
    \label{fig: example_pipeline}
\end{figure*}

Nevertheless, reconstructing neuron morphologies from microscopy images are still error-prone especially when given low-quality neuron image data. Due to the inhomogeneous fluorescence illumination and inherent light microscopy imaging limits, the raw neuron image stacks are often contaminated by numerous background noises. In addition, the voxels in dendrite and branch termini have a much lower intensity than those in soma and axon regions, which results in discontinuous neuron branches and impedes predicting the intact connections of neuron circuits. These two challenges are highlighted in the two neuron examples of Figure \ref{fig: img_to_point}. Lastly, since the 3D neuron images in the BigNeuron dataset are obtained from worldwide research laboratories and light microscopy measurements are varied, the exhibited neuron morphologies are diverse and complex. 

Various deep learning techniques have been successfully applied in medical image processing \cite{ronneberger2015u, milletari2016v, drozdzal2016importance, dou20163d, guo2015deformable}, which has inspired researchers to utilize the hierarchical feature learning ability of convolution-based models to solve the challenging neuron reconstruction problem \cite{li2017deep, wang2022ai}. In order to identify the neuron structures from a larger receptive field, recent works focus on introducing global contextual features into the convolution-based segmentation work, such as inception learning module \cite{li2017deep}, multi-scale kernel fusion \cite{wang2019multiscale}, Atrous Spatial Pyramid Pooling (ASPP) \cite{li20193d}, and graph reasoning module \cite{wang2021single}. 

In this paper, we re-think the spatial representation of neuron morphology. Rather than following the traditional 3D volumetric representation, we propose to explicitly leverage the sparsely organized point clouds to represent neuronal arbours and dendrites. As shown in Figure \ref{fig: img_to_point}, we transform the voxels of original 3D neuron images into points, then reformulate this reconstruction task to predict the geometric and topological properties for the points in cartesian coordinate system. We design a novel framework, named \emph{PointNeuron}, to extract neuronal structures from these point cloud data. Specifically, our framework consists of two major stages. The first stage is to extract a succinct neuron skeleton from the noisy input points and formulate the geometric feature. The connectivity among the unordered points is predicted at the second stage. The general idea of our method is stated in Figure \ref{fig: example_pipeline}. Our key contributions are summarized as follows: 1) we propose to describe the neuron circuits in point format for better understanding of the spatial information in 3D space, instead of original volumetric image stacks; 2) we present a novel pipeline, \emph{PointNeuron}, as an automatic 3D neuron reconstruction method by learning the characterization of point clouds, which can be generalized to improve reconstruction performance of all tracing algorithms; and 3) we present the point-based module to effectively capture the geometry information for generating the compact neuron skeletonization.

\section{Related Works}
Traditional neuron reconstruction algorithms consist of three main steps: pre-processing the raw 3D microscopy image stacks, initializing the tree-like neuronal graph map, and then pruning the reconstruction map until the compact result is obtained. APP \cite{global-peng2011automatic} and APP2 \cite{global-xiao2013app2} cover all the potential neuron signals on the raw image input for the initial reconstruction map and remove the surplus neuron branches for a compact structure at the pruning step. Like the APP family, FMST \cite{global-yang2019fmst} applies the fast marching algorithm with edge weights to initialize neuron traces and prunes them based on the coverage ratio of two intersected neuron nodes. NeuTube \cite{feng2015neutube} implements free editing functions and the multi-branch tracing algorithm from seed source points. Reversely, Rivulet \cite{zhang2016reconstruction} and Rivulet2 \cite{liu2018automated} capture the neuron traces from the furthest branch termini to the seed point. LCMBoost \cite{gu2015learning} and SmartTracing \cite{chen2015smarttracing} incorporate the deep learning-based modules into an automatic tracing algorithm without human intervention. 

With the emergence of 3D U-Net \cite{cciccek20163d} showing great success in medical image segmentation tasks, learning-based segmentation prior to applying the tracing algorithm is able to highlight the neuron signal and enhance the input neuron image quality. Some advanced deep learning techniques are applied to improve the image segmentation performance, such as inception learning module \cite{li2017deep}, multi-scale kernel fusion \cite{wang2019multiscale}, atrous convolution \cite{chen2017deeplab}, and Atrous Spatial Pyramid Pooling (ASPP) \cite{chen2018encoder, li20193d}. \cite{wang2021single, wang2022spine} incorporate graph reasoning module to the multi-scale encoder-decoder network for eliminating the semantic gap of image feature learning. For computational saving and faster inference, \cite{wang2019segmenting} proposes a light-weighted student inference model guided by the more complex teacher model via knowledge distillation. To handle the small-size neuron dataset, \cite{wang2021voxel} improves neuron image segmentation performance through the VCV-RL module extracting intra- and inter-volume voxels of same semantics into the latent space. \cite{tang20203d} builds a GAN-based framework to synthesize neuron training images from the manually annotated skeletons.

As the deep learning advances in medical image analysis, researchers have raised the interests to analyze 3D medical images by applying deep learning techniques. Although the existing works process medical images in voxel-wise representation, an increasing number of researchers are studying the 3D structures with the insight of point clouds. They leverage the 3D point cloud representation to learn more discriminative object features for different medical image tasks \cite{3dt}, such as cardiac mesh reconstruction \cite{chen2021shape}, volumetric segmentation \cite{point-unet,balsiger2019learning}, and vessel centerline extraction \cite{he2020learning}. For example, \cite{point-unet,he2020learning,balsiger2019learning} use the characterization of point clouds to learn the global context feature for enhancing the CNN-based image segmentation performance. Also, \cite{astolfi2020tractogram} and \cite{bizjak2020vascular} take into account the anatomical properties of streamline and mesh structure in the form of point cloud representation. 

The great success of introducing point cloud concepts into the domain of medical image analysis and the fact that existing tracing methods have not considered the usage of the point cloud encourage us to address the challenging neuron reconstruction task from a novel perspective. We aim to improve 3D neuron reconstruction performance through the powerful geometric and topological representation of point clouds. Therefore, we shift one of the most challenging medical image tasks to the scope of point clouds.

\section{Method}

\begin{figure*}[t]
    \begin{center}
    \includegraphics[width=\linewidth]{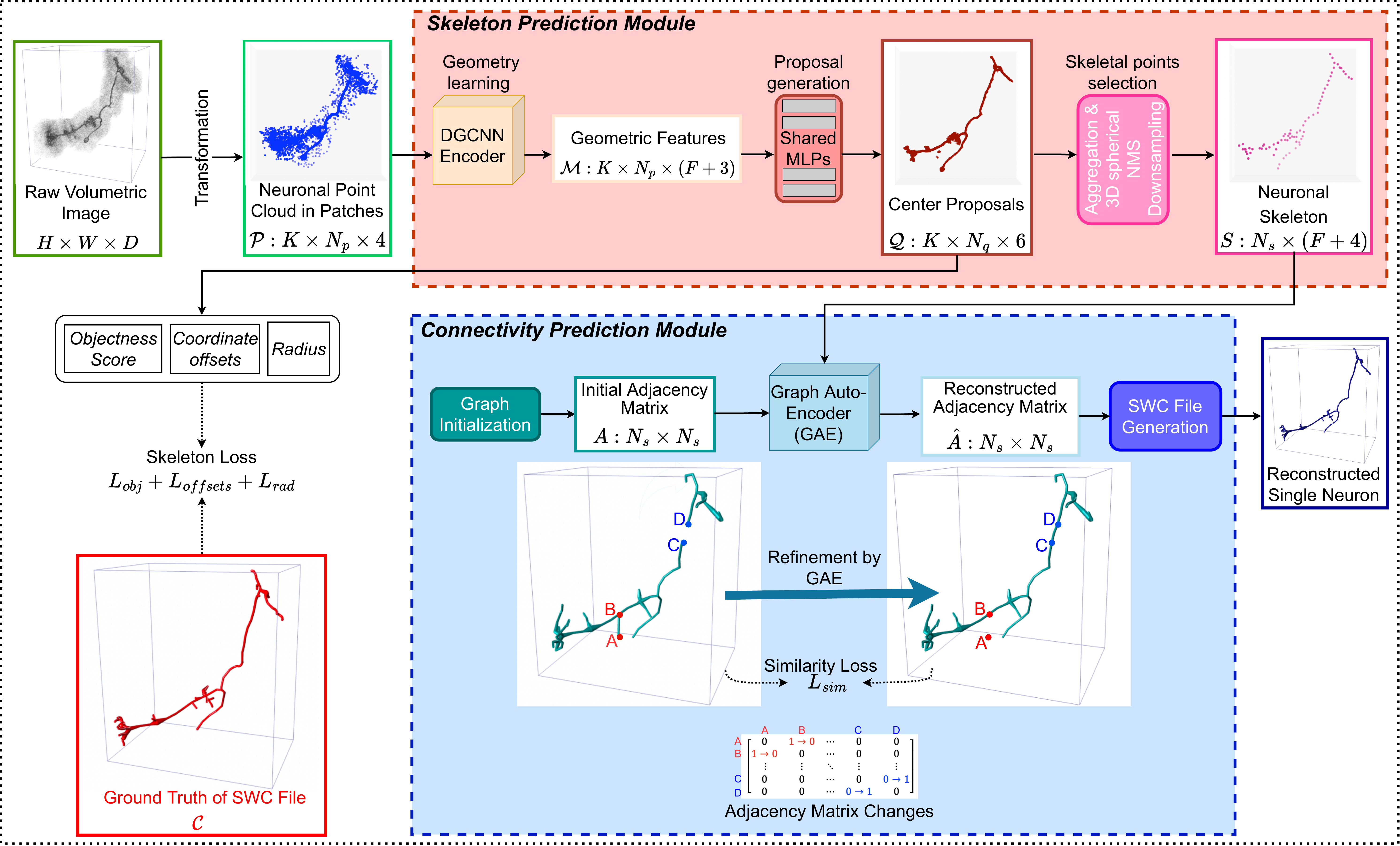}
    \end{center}
    \caption{The overview of our proposed pipeline, \emph{PointNeuron}. The neuron voxels are initially transformed to point clouds. Then a DGCNN encoder \cite{wang2019dynamic} is applied to learn deep geometric features on the neuron points. These features are processed via a proposal generation step to predict center proposals with Euclidean coordinate offsets, objectness score, and radius. To obtain a compact neuronal skeleton, we apply aggregation and 3D NMS downsampling to select the neuron skeletal points. Sequentially, we link these neuron skeletal points to produce neuron connectivity. Given the initial adjacency matrix and geometric skeleton features, we apply a graph auto-encoder to compute the correlations among points and reconstruct the adjacency matrix. After the refinement process of removing the spurs (labeled by \textcolor{red}{A} and \textcolor{red}{B} points) and bridging the gaps (labeled by \textcolor{blue}{C} and \textcolor{blue}{D} points), the final neuron reconstruction result is recorded in an SWC file. The skeleton loss and the similarity loss supervise the training process of skeleton and connectivity prediction, respectively.}
    \label{fig:framework}
\end{figure*}

We propose a novel pipeline, \emph{PointNeuron}, to perform 3D neuron morphology reconstruction in the point-based manner. Given the voxel-wise neuron image input, we initially convert it to a point cloud in Section \ref{sssec:num1}. Then we forward the neuron point cloud into the Skeleton Prediction module to generate a series of neuron skeletal points in Section \ref{sssec:num2}. After that, we design the Connectivity Prediction module to link these skeletal points through analyzing the node relationships of graph data structure in Section \ref{sssec:num3}. Lastly, we present the specific training losses in Section \ref{sssec:num4}. Our pipeline is shown in Figure \ref{fig:framework}.

\subsection{Voxel-to-Point Transformation \label{sssec:num1}}
Given a raw volumetric neuron image of size $\mathbb{R} ^{H\times W\times D}$, a thresholding value $\theta$ is pre-defined to segment the neuron structure and remove a majority of noises. Every voxel with an intensity larger than $\theta$ is positioned and transformed to a point. To handle large amount of points, we split all the neuron points into $K$ patches. Hence, the neuron point input can be represented as $\mathcal{P}=K \times \{p_{i}:[x_{i};I_{i}]\}_{i=1}^{N_{p}}$ where $N_{p}$ is the number of points per patch with the Cartesian coordinate $x_{i}\in\mathbb{R}^{3}$ and the intensity $I_{i}\in\mathbb{R}$.

\subsection{Neuron Skeleton Prediction \label{sssec:num2}}
In this module, we extract $N_{s}$ skeletal points from the neuronal point cloud input to constitute a neuron skeleton with the point-wise $F$-dimensional geometric features. There are three primary steps: learning the deep geometric features of neuron points through a graph-based encoder, generating the center proposals at local regions, and producing the compact neuron skeleton.

\noindent\textbf{Point cloud geometry learning}. 
Since the point clouds representing neuron structures are uneven and unordered in coordinate space, they cannot be simply processed by a regular grided convolution kernel like typical pixel- or voxel-wise images. Therefore, we adopt DGCNN \cite{wang2019dynamic} as our encoder to learn the spatial characteristics of neuron point clouds. The EdgeConv blocks in this architecture are designed to encode the local point-wise semantic feature, which connects a neuron point with its k-nearest neighboring points then computes and fuses edge features by graph reasoning. This local aggregation could be beneficial in complimenting topological information to neuron points. Additionally, DGCNN can yield hierarchical geometric features and incorporate multi-scale contextual information as the node relationships are dynamically refreshed after each layer. DGCNN encoder, stacking several EdgeConv blocks with residual connections, takes input points $\mathcal{P}$, and outputs the geometric features $\mathcal{M}\in\mathbb{R}^{K\times N_{p}\times (F+3)}$ where $F$ is the feature dimension. The Cartesian coordinates of points are remained the same.

\begin{figure}
  \centering
  \includegraphics[width=\linewidth]{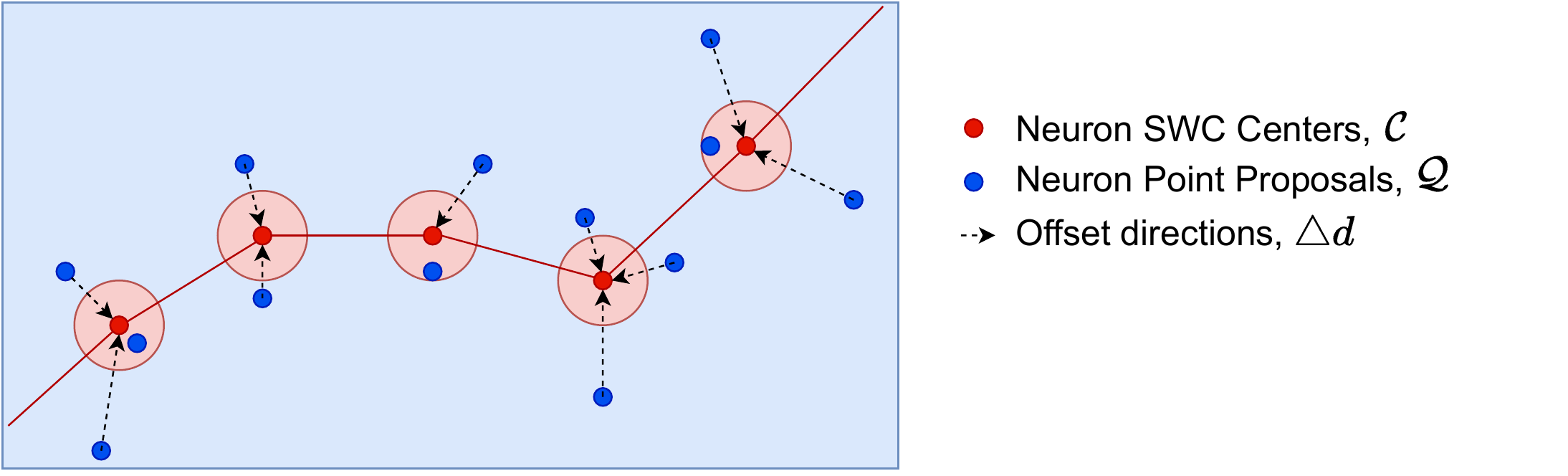}
  \caption{The surrounding points are moving toward the neuron centers denfined in the SWC files by the predicted offsets. The offsets are supervised by the distance-based losses.}
  \label{fig:point_offset}
\end{figure}

\begin{figure}[h]
    \begin{center}
    \includegraphics[width=0.9\linewidth]{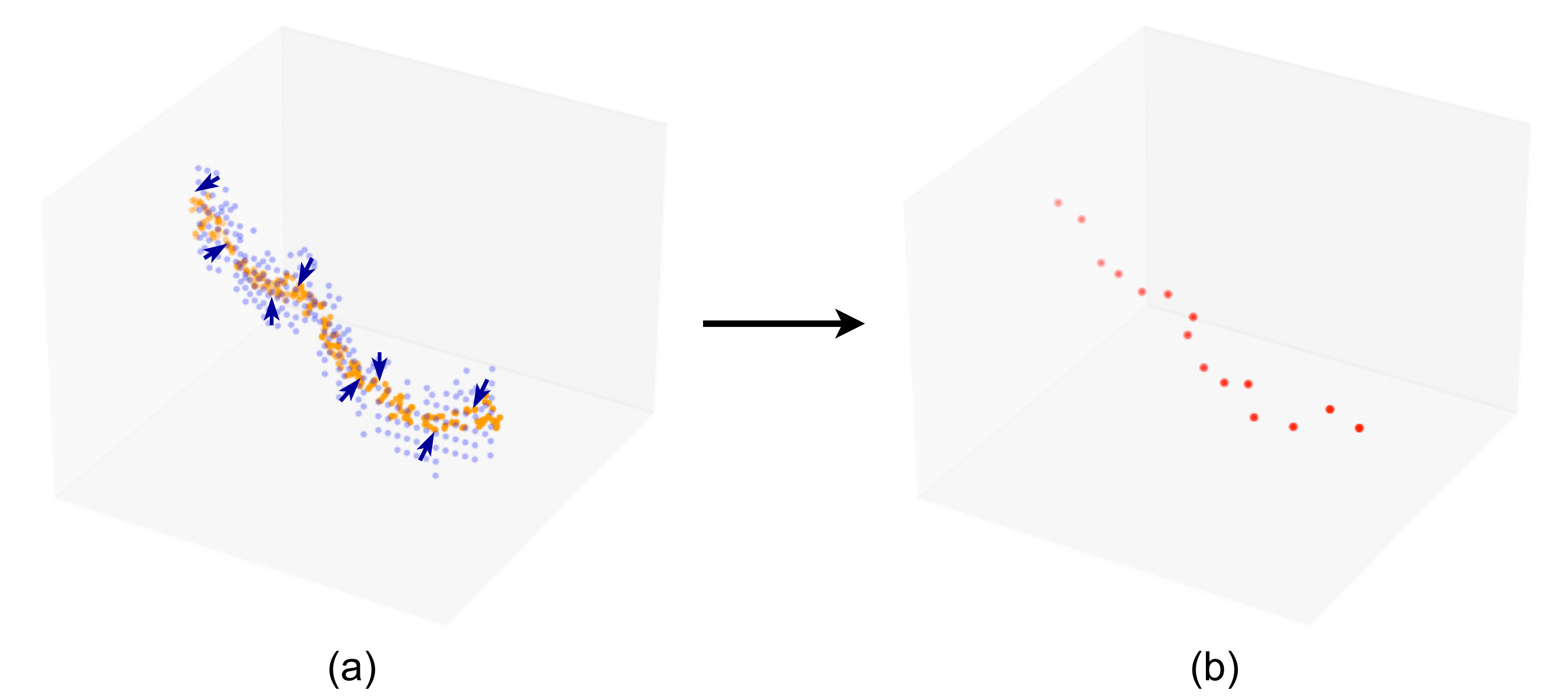}
    \end{center}
    \caption{Visualization of the neuron skeleton prediction process for a small neuron branch: (a) the raw neuron points (blue) are adaptively directed into the structure interior, then generating the preliminary neuron center proposals (orange); (b) the sampled skeletal points (red) are obtained to represent more clear neuron morphology.}
    \label{fig:interm_res_2}
\end{figure}

\noindent\textbf{Proposal generation}. 
We observe that the ground truth SWC file uses a set of key points to denote a single neuron as a hierarchical tree-like model, which constitutes a neuron skeleton, while each of key points could be considered as the center of its near neuron points at local region. The key points of SWC file are represented as $\mathcal{C}= \{c_{j}:[y_{j};r_{j}^{c}]\}_{j=1}^{N_{c}}$ where $N_{c}$ is the number of SWC key points with the coordinate $y_{j}\in\mathbb{R}^{3}$ and radius $r_{j}^{c}\in\mathbb{R}$. Inspired by this, we move the neuron point toward its nearest local center by the coordinate offsets ${\triangle d_{i}\in\mathbb{R}^{3}}$, which is schematically illustrated in Figure \ref{fig:point_offset}. It is expected that the neuron points are pushed close to the center of local region and the sparse neuron point cloud becomes more concentrated as shown in Figure \ref{fig:interm_res_2}. Consequently, we consider these moved neuron points as the center proposals $\mathcal{Q}$. In our implementation, the learned geometric features of input points are forwarded into the proposal module of a shared multi-layer perceptron (MLP). The MLP comprises of fully-connected layers, batch normalization, and LeakyReLU. The proposal module predicts the center proposals $\mathcal{Q}=\{q_i:[\tilde{x}_{i};\hat{s}_{i};\hat{r}_{i}^{q}]\}_{i=1}^{N_{q}}$ with the coordinate $\tilde{x}_{i}\in\mathbb{R}^3$, the objectness score $\hat{s}_{i}\in\mathbb{R}^2$ (we formulate the objectness score prediction as a binary classification problem supervised by a cross entropy loss), and the radius $\hat{r}_{i}^{q}\in\mathbb{R}$ are generated. The coordinate of a center proposal is obtained by adding the coordinate offsets to the input point, $\tilde{x}_{i}=x_i+\triangle d_{i}$.

\noindent\textbf{Skeletal points selection}. We aim to select the high-confidence center proposals as the neuron skeletal points from the abundant and overlapped center proposals. We thus resort to a spherical NMS dowsampling strategy with the 3D geometric information predicted by proposal generation to obtain a compact neuron skeleton $\mathcal{S}$. Non-max suppression (NMS) \cite{bodla2017soft, liu2019adaptive, girshick2014rich, girshick2015fast} is a general computer vision approach to selecting one object out of several overlapped objects. The core of this approach is to eliminate redundant object proposals through iteratively comparing the confidence score and the intersection over union (IoU) among proposals. Following this idea, we initially use the predicted Cartesian coordinates and radii of center proposals to form the potential skeletal spheres. We store the center proposals with the highest objectness score per iteration as output while discarding the remaining skeletal sphere if the IoU of spherical volume with this output is greater than a threshold $\theta_{\emph{IoU}}$. This method can effectively take the neuron skeletal points out of the center proposals. It is clearly visualized in Figure \ref{fig:interm_res_2} that these representative skeletal points constitute a succinct neuron skeleton and the neuron structural morphology is more prominently reflected than neuron input points and center proposals.

\subsection{Neuron Connectivity Prediction \label{sssec:num3}}
\begin{table*}[hb]
\caption{Quantitative comparisons with traditional algorithms for neuron reconstruction performance.}
\centering
\begin{tabular}{l|l|l|l|l|l|l}
\hline
Method                                             & ESA $\downarrow$              & DSA $\downarrow$              & PDS $\downarrow$              & Precision (\%) $\uparrow$    & Recall (\%) $\uparrow$        & F1 (\%) $\uparrow$            \\ \hline\hline
APP2                                               & $3.22_{\pm0.38}$ & $7.22_{\pm0.89}$ & $0.32_{\pm0.05}$ & $64.51_{\pm9.84}$ & $51.67_{\pm21.94}$ & $52.63_{\pm13.22}$ \\
+ 3D U-Net \cite{cciccek20163d} & $1.59_{\pm0.19}$ & $3.66_{\pm0.81}$ & $0.22_{\pm0.02}$ & $86.23_{\pm7.76}$ & $56.30_{\pm19.39}$ & $64.92_{\pm15.33}$ \\
+ 3D MKF-Net \cite{wang2019multiscale} & $1.62_{\pm0.21}$ & $3.85_{\pm0.76}$ & $0.22_{\pm0.03}$ & $89.03_{\pm8.74}$ & $55.07_{\pm18.13}$ & $65.15_{\pm1.93}$ \\
+ VCV-RL (SOTA) \cite{wang2021voxel}                                   & $1.52_{\pm0.21}$ & $\mathbf{3.48}_{\pm0.29}$ & $0.21_{\pm0.22}$ & $87.25_{\pm7.41}$ & $56.87_{\pm17.92}$ & $66.17_{\pm14.39}$ \\
+ \textbf{PointNeuron (Proposed)} & $\mathbf{1.43}_{\pm0.29}$ & $4.13_{\pm0.85}$ & $\mathbf{0.18}_{\pm0.03}$ & $\mathbf{91.56}_{\pm6.81}$ & $\mathbf{63.17}_{\pm22.85}$ & $\mathbf{71.7}_{\pm17.48}$ \\ \hline
NeuTube \cite{feng2015neutube}                                           & $2.05_{\pm0.66}$ & $5.58_{\pm1.37}$ & $0.26_{\pm0.08}$ & $87.75_{\pm8.55}$ & $52.11_{\pm15.90}$ & $62.89_{\pm12.58}$ \\
+ \textbf{PointNeuron (Proposed)} & $\mathbf{2.03}_{\pm0.87}$ & $\mathbf{5.33}_{\pm1.7}$  & $\mathbf{0.23}_{\pm0.06}$ & $\mathbf{93.13}_{\pm6.33}$ & $\mathbf{52.55}_{\pm14.9}$  & $\mathbf{65.26}_{\pm12.15}$ \\ \hline
LCMboost-FastMarching \cite{gu2015learning}                             & $1.96_{\pm0.72}$ & $4.85_{\pm1.52}$ & $0.23_{\pm0.08}$ & $85.04_{\pm9.11}$ & $44.92_{\pm18.55}$ & $55.86_{\pm16.57}$ \\
+ \textbf{PointNeuron (Proposed)} & $\mathbf{1.76}_{\pm0.63}$ & $\mathbf{4.79}_{\pm1.38}$ & $\mathbf{0.18}_{\pm0.05}$ & $\mathbf{93.31}_{\pm6.85}$ & $\mathbf{50.87}_{\pm21.35}$ & $\mathbf{63.03}_{\pm19.55}$ \\ \hline
\end{tabular}
\label{table: 1}
\end{table*}

In this section, inspired by the mesh generation of Point2Skeleton \cite{lin2021point2skeleton}, we encode the connectivity of a neuron circuit as a graph data structure. The graph is a high freedom degree structure consisting of nodes and edges. The neuron skeletal spheres are the graph nodes, and the parent-child relationship between the two points is the edge of two graph nodes.

\noindent\textbf{Graph initialization}. During training, we extract the point-wise relationships from the ground truth SWC file to generate a preliminary parent-child relationship of neuron points. The graph initialization is based on these preliminary relationships, which demonstrates basic connectivity information of the neuronal skeleton and ensures a relatively reliable topology. The reconstructed graph map is expected to be as much similar to the initialized graph as possible. Different graph initialization strategies for inference will be discussed later.

\noindent\textbf{Connectivity prediction from the reconstructed graph map}. According to this topology information, the initial undirected graph $A = (\mathcal{V}, \mathcal{E})$ is formed where $\mathcal{V}$ is the set of neuron skeletal points and $\mathcal{E}$ is the set of edges. The adjacency matrix $A$ as one input to the Graph Auto-Encoder (GAE) encodes the edge connectivity. To effectively utilize the meaningful geometry information from the skeleton prediction module, we concatenate the geometric contextual features of the neuron skeletal points $M$, the Cartesian coordinates $\tilde{x}$, and their radii $\hat{r}^{q}$ to establish the node features $X\in\mathbb{R}^{N_{s}\times(F+4)}$ where $N_{s}$ is the number of neuron skeletal points. The encoder part of GAE is a stack of graph convolution layers of graph convolution, batch normalization, and ReLU, which compresses the node feature into the latent embedding $Z$:
\begin{equation}
\label{eq:gcn}
    Z = GCN(X, A).
\end{equation}

\noindent The decoder part of GAE adopts the inner product to reconstruct the new adjacency matrix $\hat{A}$:
\begin{equation}
\label{eq:gcn}
    \hat{A} = Z\cdot Z^{T}.
\end{equation}

\noindent\textbf{SWC file generation}. Finally, we design a recursive method to interpret the predictions of neuron skeletal points and the reconstructed connectivity. It can recursively search the detected neuron points and edges to rebuild the target SWC file.

\subsection{Learning Objectives \label{sssec:num4}}
For neuron skeleton prediction, we expect that our predicted skeleton reflects the identical structural meaning as the key points manually annotated in the SWC file.

\noindent\textbf{Coordinate offset loss}. The bi-directional Chamfer Distance (CD) is measured between the SWC key points and center proposals to optimize the predicted coordinate offsets: 
\begin{equation}
\label{eq:center_loss}
	L_{\emph{offsets}} = \sum_{\scriptscriptstyle a \in \{y_{j}\}} \min_{\scriptscriptstyle b \in \{\tilde{x}_{i}\}} \Vert a - b \Vert_{2}^{2}+ \sum_{\scriptscriptstyle b \in \{\tilde{x}_{i}\}} \min_{\scriptscriptstyle a \in \{y_{j}\}} \Vert b - a \Vert_{2}^{2}.
\end{equation}

\noindent\textbf{Objectness loss}. In addition to explicilty directing the points, we introduce a cross-entropy loss to supervise the predicted objectness score indicating how confident a local center proposal locates inside of the neurite region:
\begin{equation}
	L_{obj} = -\frac{1}{N_{p}} \sum_{i} s_{i} \log(\sigma(\hat{s}_{i})) + (1-s_{i}) \log(1-\sigma(\hat{s}_{i})),
\end{equation}
where $\hat{s}_{i}$ is the objectness score prediction, $s_{i}$ is the objectness label, and $\sigma$ is the Sigmoid activation function. We assign the center proposals entered into the ground truth SWC sphere to have true objectness labels and the others to have false objectness labels. These two complemental loss functions encourage the consistency of the predicted neuron center proposals and ground truth SWC points from two perspectives of Euclidean geometry and object confidence, which are discussed in Section \ref{section: ablation} - Ablation Study.

\noindent\textbf{Radius loss}. The radius of a neuron point is supposed to be the same as the radius of the nearest center. We use Mean Absolute Error (MAE) to calculate the radius loss:
\begin{equation}
\label{eq:radius_loss}
	L_{rad} = \frac{1}{N_{p}}\sum_{i} \vert \xi(\mathcal{C}, q_{i}) - \hat{r}_{i}^{q}\vert,
\end{equation}
where $\xi(\mathcal{C}, q_{i})$ represents the radius of the nearest SWC point to the center proposal $q_{i}$.

These three losses are summed with weight $\lambda$ to supervise the skeleton prediction module:
\begin{equation}
    L_{skel} = L_{\emph{offsets}} + \lambda L_{obj} + L_{rad}.
\end{equation}

When training the neuron connectivity prediction module, we aim to make the reconstructed adjacency matrix as similar as possible to the input adjacency matrix initialized by the SWC file.

\noindent\textbf{Similarity loss}. We apply a masked cross entropy loss to supervise GAE learning:
\begin{equation}
L_{sim} = \frac{1}{N_{s}} (M \odot ( A\log(\sigma(\hat{A})) -  (1-A)\log(\sigma(1-\hat{A})))),\\ 
\end{equation}
where $M$ is the mask of the interested points, and $\sigma$ is the Sigmoid activation function.

\section{Experiments and Results}
\begin{figure*}[hb]
    \centering
    \includegraphics[width=0.9\linewidth]{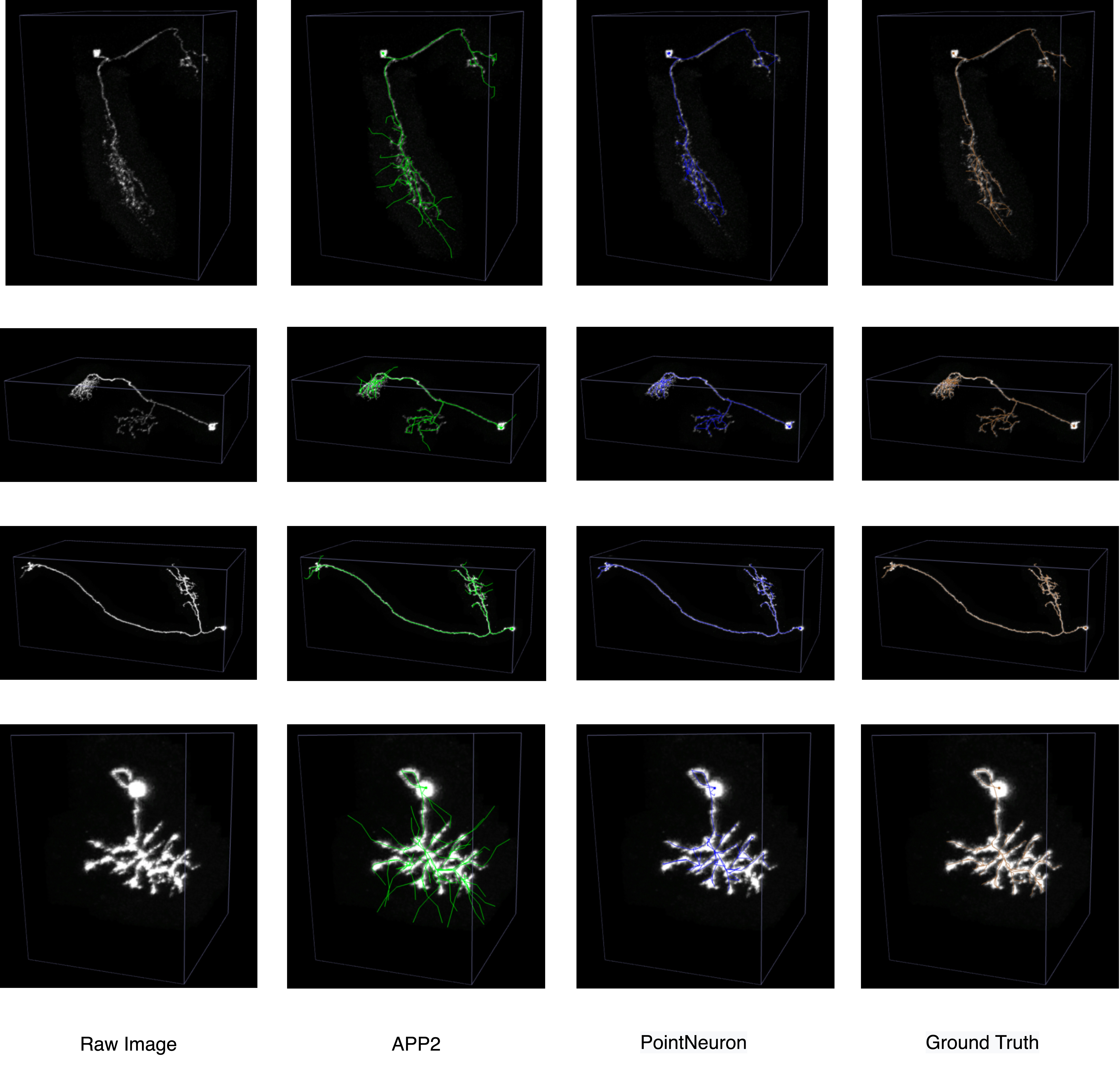}
    \caption{Visualization of the neuron tracing performance comparisons on test. Each row represents one image.}
    \label{fig:vis_app2}
\end{figure*}

\subsection{Dataset and Implementation Details}
\noindent\textbf{Dataset}. Our proposed framework is evaluated on the Janelia-Fly dataset of the BigNeuron project, which features diverse neuron morphologies of the fruitfly on cortical and subcortical areas, retina, and peripheral nervous system regions \cite{manubens2022bigneuron}. The volumetric images of this dataset were imaged from light microscopy, and their manual annotations were processed by the Brainbow method \cite{livet2007transgenic}. We separate the total 42 images into 35 training images, 3 validation images, and 4 testing images. The average resolutions for training and testing images are $197\times198\times168$ and $262\times159\times181$. The foreground neuron voxels are segmented out by intensity thresholding (of $\theta=0.2$). By mapping these foreground voxels into points in Cartesian space, we obtain around $N=12000$ points per image. The number of points per image in this task is much larger than the number of point clouds for normal instances in ModelNet40 \cite{wu20153d} and ShapeNet \cite{chang2015shapenet}. A point cloud patch of $N_{p}=512$ is randomly cropped from the entire neuronal points as the input points for training our proposed framework, and data augmentation is applied including rotation and flipping.

\noindent\textbf{Network architecture details}. We stack 3 EdgeConv blocks in DGCNN to encode the geometric contextual information, then the geometric features are processed by the proposal module of 3 shared MLP layers to generate center proposals. For connectivity prediction, GAE of 12 layers with residual learning is used. More details are provided in the supplementary material.

\noindent\textbf{Training the network}. The neuron skeleton prediction and neuron connectivity prediction modules are separately trained with Adam optimizer. For skeleton prediction module, the backbone DGCNN network and proposal module are employed together to generate center proposals, which are trained with $L_{skel}$ (with $\lambda=10$) and a learning rate $1e^{-3}$. 3D spherical NMS with the acceptable range of $\theta_{\emph{IoU}}$ is from 0.05 to 0.25 is to select high-confidence center proposals for obtaining a compact neuron skeleton. When training connectivity prediction module, we freeze the skeleton prediction module and train the GAE with $L_{sim}$ and a learning rate of $5e^{-4}$. Training these two modules to convergence on GeForce RTX 2080 Ti takes 1200 and 200 epochs, respectively.


\noindent\textbf{Inference}. Our proposed skeleton prediction module predicts partial neuron skeleton from a window sliding along the neuron structure. Then, these outputs are aggregated to formulate the complete skeleton. Afterwards, linking the neuron skeletal nodes explores the neuronal relationships and outputs the final neuron reconstruction as a new SWC file in connectivity prediction module. 

\subsection{Results and Analysis}
We use the three spatial distance-driven metrics proposed by \cite{peng2010automatic}, Entie Structure Average (ESA), Different Structure Average (DSA), and Percentage of Different Structure (PDS), to measure the geometric discrepancy between the neuron reconstruction and the ground truth SWC file. The calculation of these three metrics takes into account the correctness of connectivity of neuron nodes, which is processed by the Vaa3D software plugin. We report the reconstruction performance using Precision, Recall, and F1, where true positive (TP), false postive (FP), and false negative (FN) are determined by measuring how far a predicted neuron point is away from the nearest ground truth points. 

The quantitative evaluations of our proposed method and traditional tracing algorithms are shown in Table \ref{table: 1}. Aiming to facilitate comparisons with traditional algorithms and emphasize the performance improvements from our method, ``+" means that the previous neuron tracing algorithms (APP2, NeuTube, and  LCMboots) were used for graph initialization of the connectivity prediction module of our proposed framework. As indicated in Table \ref{table: 1}, the tracing performance with our proposed method is better than the traditional algorithms. Based on the same APP2, our proposed method outperforms the previous deep voxel-based models and is comparable to the SOTA work of VCV-RL \cite{wang2021voxel}. We present a visualization comparison of reconstructions with APP2 and ours in Figure \ref{fig:vis_app2}. The predicted neuron structures achieve a better match to the ground truth spatially, and most neuron skeletal nodes are considered to be structurally meaningful. It is noticeable that the reconstruction results from our method produce less false positive points (or branches) than APP2 and are closer to the ground truth. 

\subsection{Model Complexity}
We measure the complexity of our proposed point-based model with comparisons to the previous deep learning voxel-based models. Table \ref{tabel:3} shows the remarkable advantages of our point-based model in terms of computational memory take-up and inference speed. With the input data type as point cloud, we eliminate the computing memory need for the majority of senseless background voxels on a volumetric microscopy image and reduce the input memory size by 95.6\%. In addition, our proposed MLP-dominated model requires fewer trainable parameters than convolution-based networks, while being much more efficient in the inference procedure.

\begin{table}[h]
\centering
\caption{Model complexity analysis of our point-based model with previous voxel-based models for one pass inference.}
\resizebox{\linewidth}{!}{
\begin{tabular}{l|c|c|c}
\hline
Model             & \multicolumn{1}{l|}{Input memory size (MB)} & \multicolumn{1}{l|}{Params (M)} & \multicolumn{1}{l}{Inference time (ms)} \\ \hline
3D U-Net          & 33.55                                       & 1.50                            & 917.38                                  \\ \hline
3D MKF-Net        & 33.55                                       & 1.55                            & 925.69                                  \\ \hline
VCV-RL            & 33.55                                       & 1.53                            & 1038.67                                       \\ \hline
\textbf{Proposed} & \textbf{1.47}                               & \textbf{1.17}                   & \textbf{665.34}                         \\ \hline
\end{tabular}
}
\label{tabel:3}
\end{table}

\subsection{Ablation Study} \label{section: ablation}
We conduct an ablation study as listed in Table \ref{table:2}. Models 1 and 2 are configured with different losses, demonstrating that the offsets loss $L_{\emph{offsets}}$ effectively pushes the neuron points inward, but a single geometric loss cannot reach the optimal result. In Model 3, the encoder in skeleton prediction is replaced with another point-based backbone, PointNet++, and it does not perform as efficently as DGCNN, given that DGCNN employs a dynamic design to construct the graph map in each layer which is superior for learning global representative features in latent space. We also attempt to directly forward the plain point feature of Cartesian coordinate and intensity into the GAE in the connectivity prediction module, and the results of Model 4 illustrate that the geometric information of neuronal points learned from the skeleton prediction module is essential to generating accurate links between neuron points. Furthermore, we test different downsampling methods. Our proposed spherical NMS downsampling outperforms the regular FPS and uniform sampling. The reason is that it takes into account the local geometric structure, and its principle is suited to handling the unevenness of the point cloud.

\begin{table}[ht]
\centering
\caption{Quantitative ablation studies with different configurations.}
\resizebox{\linewidth}{!}{
\begin{tabular}{cl|l|l|l|l}
\hline
\textbf{ID} & Method						   		& ESA$\downarrow$                   & DSA$\downarrow$                   & PDS$\downarrow$                   & F1 score (\%) $\uparrow$       \\ \hline
\textbf{1}  & w/o $L_{obj}$                      		 &$1.61_{\pm0.20}$           &$4.71_{\pm0.73}$            &$0.20_{\pm0.03}$            &$70.37_{\pm17.87}$                      \\ \hline
\textbf{2}  & w/o $L_{\emph{offsets}}$              		   &$1.70_{\pm0.17}$           &$4.74_{\pm0.61}$            &$0.22_{\pm0.2}$             &$70.83_{\pm18.17}$                      \\ \hline
\textbf{3}  & PointNet++        		   		   &$1.50_{\pm0.33}$           &$4.28_{\pm0.99}$            &$0.19_{\pm0.04}$          &$70.96_{\pm17.76}$     \\ \hline
\textbf{4}  & w/o DGCNN         	  		  &$5.06_{\pm2.07}$         &$6.89_{\pm2.36}$          &$0.38_{\pm0.06}$          &$58_{\pm34.32}$                      \\ \hline
\textbf{5}  & FPS downsampling  		   &$1.46_{\pm0.31}$           &$4.15_{\pm0.73}$            &$0.19_{\pm0.04}$           &$68.67_{\pm18.84}$                      \\ \hline
\textbf{6}  & Uniform dowsampling    &$1.51_{\pm0.28}$           &$4.19_{\pm0.67}$             &$0.20_{\pm0.03}$           &$68.65_{\pm18.83}$                    \\ \hline
\textbf{7}  & Proposed                            &$\textbf{1.43}_{\pm0.29}$          &$\textbf{4.13}_{\pm0.85}$              &$\textbf{0.18}_{\pm0.03}$           &$\textbf{71.7}_{\pm17.48}$                     \\ \hline
\end{tabular}
}
\label{table:2}
\end{table}

\section{Conclusion}
In this paper, we propose to represent neuron structures with point cloud data and design a novel point-based pipeline, \emph{PointNeuron}, to solve the challenging problem of automatic 3D neuron reconstruction. Taking transformed neuron point clouds as input, a skeleton prediction module is designed to obtain the neuronal skeletons, after which a connectivity prediction module learns the linkage among skeletal points and produce the final reconstructions. Experiments show that our point-based approach is more efficient and able to achieve finer reconstruction results compared to those based on traditional 3D volumetric data, which should facilitate 3D medical analysis and point-cloud-driven healthcare applications.

\newpage
{\small
\bibliographystyle{ieee_fullname}
\bibliography{egbib}
}

\clearpage
\twocolumn[
\begin{@twocolumnfalse}
\section*{\centering{PointNeuron: 3D Neuron Reconstruction via Geometry and Topology Learning of Point Clouds \\ Supplementary Material}}
\end{@twocolumnfalse}
]

\setcounter{figure}{0}
\begin{figure*}[hb]
    \begin{center}
    \includegraphics[width=0.95\linewidth]{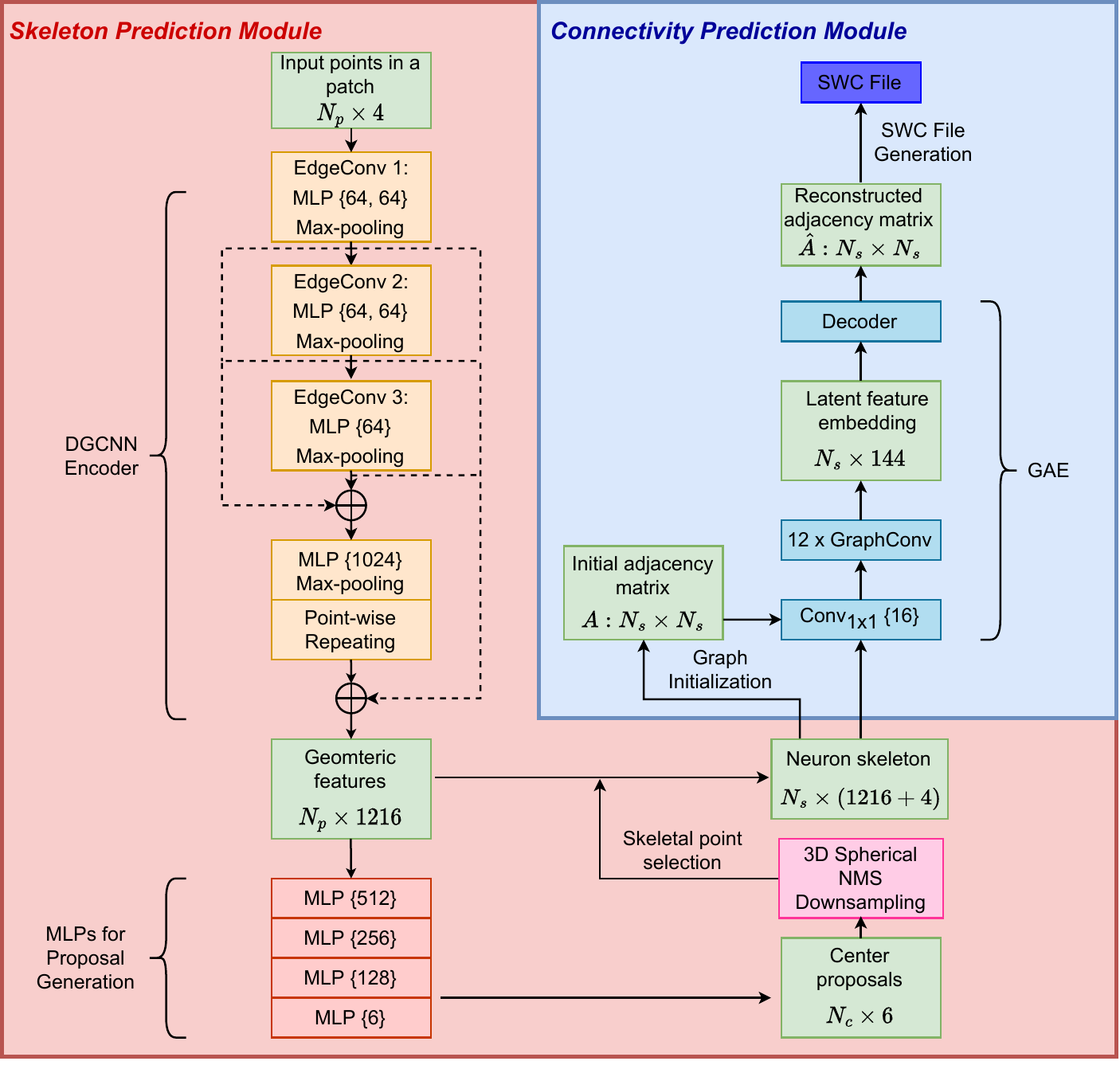}
    \end{center}
    \caption{Overall \emph{PointNeuron} architecture. $\bigoplus$ means the channel-wise concatenation.}
    \label{fig:1}
\end{figure*}

\section*{Network Architecture Details}

As mentioned in the main paper, our proposed pipeline for the 3D neuron reconstruction task, \emph{PointNeuron}, is composed of two major modules: a skeleton prediction module and a connectivity prediction module. Figure \ref{fig:1} shows the details of our designed network architecture.

The encoder for the neuron point cloud input, based on DGCNN, has 3 EdgeConv blocks. Each EdgeConv block is used to compute edge features for each point with 20 nearest neighboring points, which has a MLP network $\{c_{1},..,c_{n}\}$ where $c_{i}$ is the number of output channels of the layer $i$, and a max-pooling layer to generate the locality-aware feature maps. Each MLP layer is followed by a batch normalization and a LeakyReLU activation function with a negative slope of 0.2. These feature maps with multi-scaled receptive fields are concatenated as one through residual connections. As per the main paper, the proposal generation is based on the shared MLP layers to process geometric features, which outputs the center proposals of a 6-channel depth dimension that contains two objectness score values, a radius value, and three XYZ coordinate offset values. The last MLP layer in the proposal generation is not followed by a batch normalization and a ReLU activation function. After 3D spherical NMS downsampling, we could select the skeletal points (or sampled critical center proposals) to constitute the compact neuron skeleton.

The geometric features of skeletal points are concatenated with the point geometric predictions of Cartesian coordinates and radii, which are used as the graph node features in the connectivity prediction module to produce the relationships among skeletal points. The graph adjacency matrix is initialized by a ground truth SWC file when training, or by a traditional tracing algorithm when inferring. Each GraphConv is followed by a batch normalization and a ReLU activation function, and the output feature channels of 12 GraphConv blocks in GAE are $\{32,32,48,64,64,80,96,96,102,128,128,144\}$ to compress the node features into the latent embedding. The reconstructed adjacency matrix is obtained through the inner product decoder. In the end, based on the predicted neuron geometric skeleton information and point-wise topology relationships, we design a recursive function to generate the target SWC file.

\end{document}